\documentclass{article}

\usepackage[final]{nips_2018}

\usepackage[utf8]{inputenc} 
\usepackage[T1]{fontenc}    
\usepackage{hyperref}       
\usepackage{url}            
\usepackage{booktabs}       
\usepackage{amsfonts}       
\usepackage{nicefrac}       
\usepackage{microtype}      
\usepackage{color}
\usepackage{graphicx}
\usepackage{subfig}

\title{Bad practices in evaluation methodology relevant to class-imbalanced problems}

\author{
Jan Brabec\textsuperscript{1,2}, Lukas Machlica\textsuperscript{1}\\
\textsuperscript{1}Cisco Systems, Inc., Charles Square Center, Karlovo Namesti 10, Prague, 12000\\
\textsuperscript{2}Dept. of Computer Science, FEE, CTU in Prague, Karlovo Namesti 13, Prague, 12000\\
janbrabe@cisco.com, lumachli@cisco.com
}

\begin{document}

\maketitle

\begin{abstract}
For research to go in the right direction, it is essential to be able to compare and quantify performance of different algorithms focused on the same problem. Choosing a suitable evaluation metric requires deep understanding of the pursued task along with all of its characteristics. We argue that in the case of applied machine learning, proper evaluation metric is the basic building block that should be in the spotlight and put under thorough examination. Here, we address tasks with class imbalance, in which the class of interest is the one with much lower number of samples. We encountered non-insignificant amount of recent papers, in which improper evaluation methods are used, borrowed mainly from the field of balanced problems. Such bad practices may heavily bias the results in favour of inappropriate algorithms and give false expectations of the state of the field.
\end{abstract}

\section{Introduction}

Many problems in machine learning are inherently imbalanced in the sense that some of the classes of interest are significantly less prevalent than the other (background) classes. In medicine, only a minority of population suffers from a particular disease. In network security, most of the network traffic is benign. Other notable examples are fraud detection or disaster management, even object detection in an image can be regarded as an imbalanced task since the object of interest often occupies only a small region of the image. Generally, many fields are concerned with rare events and their detection \cite{imbalanced_review}.

Because of the high skew toward the majority class, suitable evaluation methods have to be chosen in order to understand the merit of the pursued algorithms. Evaluation techniques well suited for imbalanced problems are already known, they have been described in the past \cite{he2009learning}, but they are still often not applied correctly or misinterpreted. 

Using accuracy or classification error as an evaluation metric in the imbalanced scenario is generally known to be a bad practice \cite{chawla2009data}, but it is still very prevalent. The problem with accuracy is that it does not provide any information on the distribution of errors between individual classes, and in the presence of a high imbalance ratio, a trivial classifier that predicts only the majority class all the time will achieve high accuracy (e.g. for imbalance ratio 1:1000, we would get accuracy of 99.9\%). In \cite{imbalanced_review} authors evaluated 517 papers concerned with imbalanced classification across multiple domains. They found 201 (38\%) out of those papers to be using accuracy as an evaluation metric. However, authors of all those 517 papers knew that they are dealing with an imbalanced problem. There is a large portion of papers in which the authors are not aware of it. Most of these papers will probably tend to use accuracy based evaluation metrics as well. We have even encountered several papers where the accuracy of the proposed method is lower than the imbalance ratio. Such method would be therefore outperformed by a trivial classifier predicting the majority class only. If the problem is imbalanced, but it is not explicitly stated in the paper, accuracy of 97\% might seem perfectly fine even though the class imbalance ratio is 1:100 yielding 99\% accuracy for the trivial classifier.

In this paper we focus on several bad practices when handling imbalanced data sets that are often presented in the latest machine learning publications. After consideration, we decided not to point at specific papers with bad practices involved. To mention only several papers would not be fair to their authors considering that we are not able to perform an exhaustive search and many other papers would still be left out. We were able to identify such works among top conferences related for example to medical imaging in radiology \cite{RSNA}, intrusion detection in network security \cite{CCS}, or in the field of biometrics such as speaker recognition \cite{Interspeech2017}. Among others, this may be attributed to the gap between commercial products and academic focus of the presented work. Not all academic research has to be related to a commercial product, therefore assessing working points at which the solution could be deployed is often not the main concern. We argue that in applied machine learning, proper evaluation schemes are essential and should be in the spotlight of any review process.

In the following, we review the most common evaluation methods and consequently discuss different bad practices in relation to the class imbalance problem.


\section{Performance curves}
Common performance curves used in the literature are Receiver Operating Characteristic (ROC) and Detection Error Tradeoff (DET) curve. The latter is used dominantly in biometric systems, e.g. in speaker recognition~\cite{SpeakerVer2017}. ROC depicts the relationship between True Positive Rate (TPR), TPR = $\frac{TP}{TP+FN}$ and False Positive Rate (FPR), FPR = $\frac{FP}{FP+TN}$ \cite{he2009learning}, while DET curve shows the dependency of False Rejection Rate (FRR), FRR = 1-TPR on FPR \cite{Martin97DET}.

Important property of ROC or DET curve is that it does not take the class priors into account. Therefore, uniformly sub-sampling any of the classes (i.e. changing the class priors) in the evaluation data set will not alter the curve. This property is one of the strengths of the ROC curve, because performance of a classifier can be assessed even when the prior changes. For example, in network security traffic volume related to a specific malware may strongly vary in time. This will cause differences between priors in training and test sets. 

Precision = $\frac{TP}{FP+TP}$ and recall = $\frac{TP}{TP+FN}$ (note that this is equal to TPR) assume that only one of the classes, most often denoted as the positive class, is of interest. Precision, defined as the number of correct detections divided by the number of all detections, represents therefore the probability that the detection of the system will be correct. As will be discussed later, this cannot be inferred from the ROC/DET curve only. From this point of view, Precision-Recall (PR) curve is a better evaluation metric in scenarios where the positive predictions of the system are presented directly to the end user who has limited resources to process them. Examples of such scenarios include diagnosis presented to a doctor or potential security incidents shown to a human analyst. These scenarios also have in common that the end user does not care about the negative class.

\section{Disregarding true class priors}

True class priors are sometimes known or can be estimated. However, very few papers actually use or discuss their values even though they are necessary in order to determine if the proposed method could be used in practice.

\subsection{Performance curves}
 
In applied papers, \textit{performance curves should always be accompanied with a discussion of imbalance ratios that are encountered in the wild}. For example, in tasks with Imbalance Ratios (IRs) of 0.01, it is reasonable to study false-positive regions lower than $0.01$. Note that if we had TPR equal to one at the point FPR = 0.01 with IR = 0.01, the true precision at this point would be TPR/(FPR/IR + TPR) = 0.5, that is 50 out of 100 hundred detections would be false alarms. In domains with high IRs, such as network security or detection of very rare diseases, such performance would be inadequate.  Also note that \textit{it is impossible to compute the precision from ROC curve unless the imbalance ratio is given}. This means that investigation of ROC curves without the knowledge of IR will not answer the question whether the system in question could be used in practice, i.e. not produce too many false alarms presented to the user.

We suggest to always discuss possible imbalance ratios and if reasonable (see Section \ref{sec:precision}) report also Precision-Recall (PR) curves along with the ROC/DET curve. This is useful especially when positive predictions of the system are presented to the end user with limited resources (e.g. time) to process them. Precision will directly answer the question what percentage of detections are going to be truly positive. 

Using PR curves alone may be seen as not considering the performance of the system on the negative class. However, this information may still be of significant importance. For example, when selecting passengers for detailed inspection at airport security, we wish to know what the percentage of affected passengers is. Another example may be related to costs of drugs provided to patients identified by an algorithm: what percentage of healthy users will be affected by the experiments? In intrusion detection, especially in blocking technologies, the main point to be investigated is how many clean devices the system will affect.

\subsection{Precision}
\label{sec:precision}
When experimental data sets are created and the minority class is rare, often all the available samples from this class are kept, while down-sampling the majority class. This may be reasonable in training but not in testing since the true imbalance ratio, which the classifier will encounter once deployed, is violated.
In this situation, using precision computed directly on the test set is misleading. \textit{Assuming that the imbalance ratio is smaller in the test data set than in the wild, reported precision is overly optimistic, because the number of false positives will be higher in real setting.} The same holds for F-score, which is the harmonic mean of precision and recall, often used as a single number depicting both precision as well as recall. Therefore, we should strive for the test set to resemble true data distributions as much as possible. Ideally without any sampling at all.

In some cases we are not able to reproduce the true imbalance ratio in our test data set without down-sampling the minority class which is in most of the cases not desirable. For example, if the prevalence of a disease in population is 1:$10^4$, we would need to obtain $10^4$ labeled negative samples for the test data set per each positive sample, which is often not achievable.

ROC curve (with imbalance ratios discussed) is preferred in these cases as it is independent of the class priors. Another possibility is to adjust the precision to take the real imbalance ratio into consideration. Consider a binary classification problem where the prior of the minority class on the test set is $p_{test}$ while the real prior is $p_{real}$. We can estimate the precision for the minority class for $p_{real}$ given only the test set according to:

\label{adjusted_precision}
$$
adjusted\_precision =  \frac{\frac{p_{real}}{p_{test}} \cdot \mathrm{TP}}{\frac{p_{real}}{p_{test}} \cdot \mathrm{TP} + \frac{1-p_{real}}{1-p_{test}} \cdot \mathrm{FP}} \cdot
$$
Adjusted precision is a linear rational function of $p_{real}$, and as such can be plotted (preferably with a logarithmic scale of the x-axis) to show how the classifier's precision depends on the imbalance ratio. The discussion of this effect on the precision of the proposed classifier is of interest for practical applications. 

Similarly, F-score can be adjusted using the adjusted precision.

\section{Area Under Curve (AUC) and regions of interest}

The area under ROC curve is commonly used to compare the performance of two classifiers in the form of a single number. The issue occurs when AUC includes regions that are of no interest, because the classifier would never be used at those operating points. \textit{In the case of imbalanced data sets, the regions of no interest may represent most of the area under the curve, having dominant influence on the value of AUC.}

\begin{figure}%
    \centering
    \subfloat{{\includegraphics[width=0.48\linewidth]{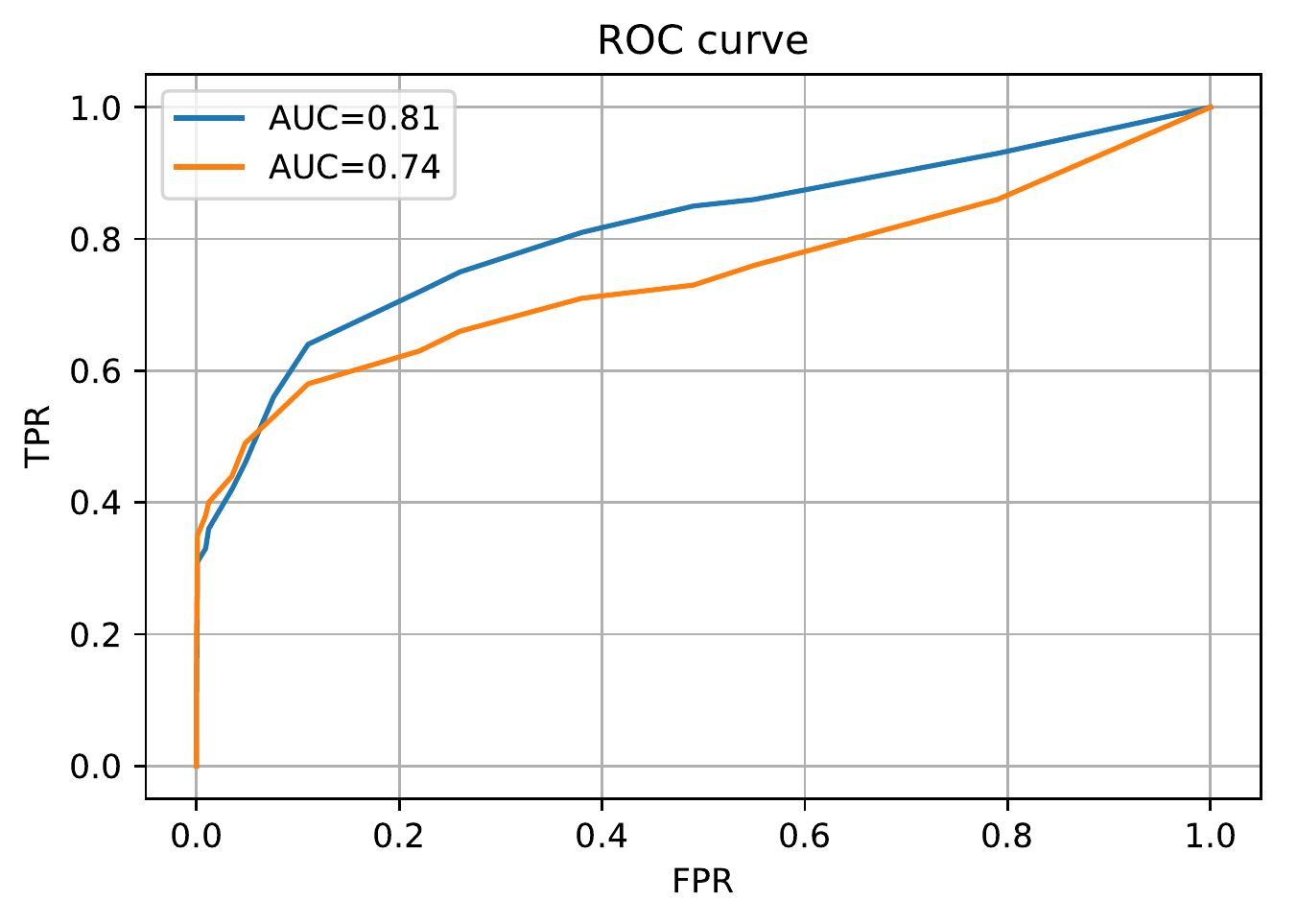} }}%
    \subfloat{{\includegraphics[width=0.49\linewidth]{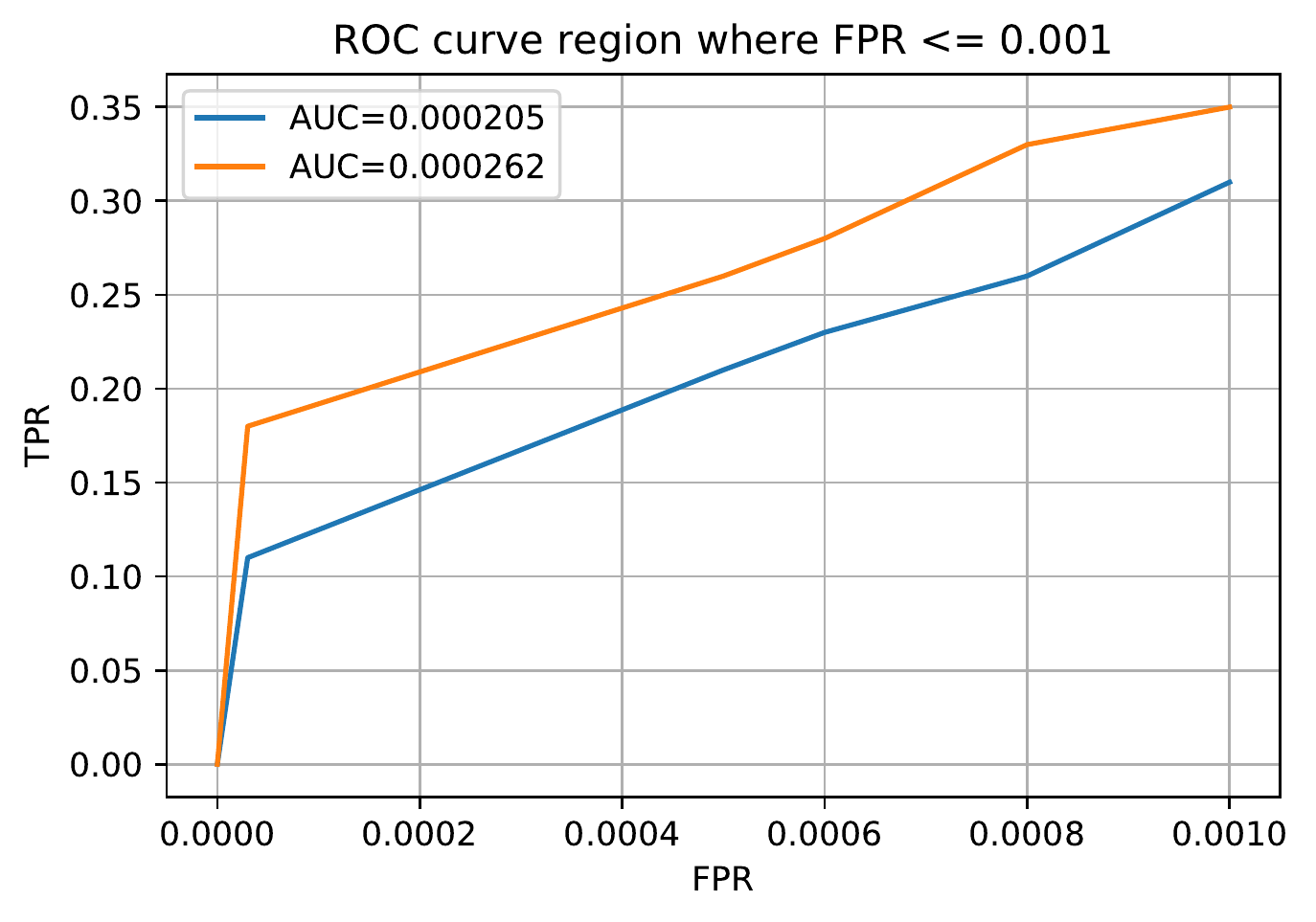} }}%
    \caption{Example of two ROC curves plotted in different regions of FPR with imbalance ratio 1:$10^3$. While in the left figure the AUC is 5\% higher for the blue curve, the orange curve has greater AUC in the region of interest (right figure). Any drops in performance in regions for FPR higher than $10^{-3}$ can be treated as noise in the final outcome.}%
    \label{fig:auc_example}%
\end{figure}

For example, in the domain of network traffic intrusion-detection, the imbalance ratio is often higher than $1:10^{3}$ \cite{garcia2014empirical}, and the cost of a false alarm for an applied system is very high. This is due to increased analysis and remediation costs of infected devices. In such systems, the region of interest on the ROC curve is for false positive rate at most $10^{-4}$. \textit{If AUC was computed in the usual way over the complete ROC curve then $99.99\%$ of the area would be irrelevant and would represent only noise in the final outcome.} We demonstrate this phenomenon in Figure \ref{fig:auc_example}.

If AUC has to be used, we suggest to discuss the region of interest, and eventually compute the area only at this region. This is even more important if ROC curves are not presented, but only AUCs of the compared algorithms are reported.

\section{Concluding remarks}

Proper evaluation methodology is necessary to obtain the right insights and make correct decisions. We do not regard any of the metrics described herein as a silver bullet but argue for a proper use of accuracy, precision, ROC or PR curves based on the identified imbalance ratio and the structure of the test set. Note that several weighted variants of accuracy, ROC, AUC exists and were defined for imbalanced scenarios, but they are not wide-spread and in the majority of cases standard variants are used.

\def\bibfont{\footnotesize}
\setcitestyle{numbers}
\bibliographystyle{plain}
\bibliography{imbalanced}

\end{document}